\documentclass[sigconf]{acmart}
\usepackage{textcomp}
\usepackage{subcaption}
\usepackage[font=small,skip=0pt]{caption}

\usepackage{float}

\usepackage{acronym}

	\newacro{MT} {Machine Translation}
	\newacro{SMT}{Statistical Machine Translation}
	\newacro{NMT}{Neural Machine Translation}
	\newacro{RNN}{Recurrent Neural Network}
	\newacro{LSTM} {Long Short Term Memory}
	\newacro{BLEU}{Bilingual Evaluation Understudy Score}
\newacro{NLTK}{Natural Language Translation Kit}
	\newacro{GPU}{Graphical Processing Unit}	
	\newacro{GRU} {Gated Recurrent Unit}
	\newacro{WFST}{Weighted Finite State Transducer}
	\newacro{HMM}{Hidden Markov Model}  
	\newacro{SGD} {Stocastic Gradient Descent}
	\newacro{POS} {Part Of Speech}
	
\newacro{English to Malay} {EN $\rightarrow$ ML}
\newacro{Malay to English} {ML $\rightarrow$ EN}
	\newacro{French to English} {FR$\rightarrow$ EN}
	\newacro{English to French} {EN $\rightarrow$ FR}

\newcommand{\EM} { EN $\rightarrow$ ML }
\newcommand{\ME} { ML $\rightarrow$ EN }

\AtBeginDocument{%
  \providecommand\BibTeX{{%
    \normalfont B\kern-0.5em{\scshape i\kern-0.25em b}\kern-0.8em\TeX}}}

\setcopyright{acmcopyright}
\copyrightyear{2020}
\acmYear{2020}
\acmDOI{10.1145/1122445.1122456}

\acmConference[to be published by ICNLP 2020]{International Conference on Natural Language Processing}{July 11-13, 2020, submitted version}{}
\acmBooktitle{ICNLP 2020: International Conference on Natural Language Processing, July 11-13, 2020}

\begin{document}

\title{Neural Machine Translation model for University Email Application}


\author{Sandhya Aneja}
\affiliation{\institution{Faculty of Integrated Technologies\\Universiti Brunei Darussalam\\Brunei Darussalam}}
\email{sandhya.aneja@ubd.edu.bn}

\author{Siti Nur Afikah Bte Abdul Mazid}
\affiliation{\institution{Faculty of Integrated Technologies\\Universiti Brunei Darussalam\\Brunei Darussalam}}
\email{16b4005@ubd.edu.bn}

\author{Nagender Aneja}
\affiliation{\institution{Institute of Applied Data Analytics\\Universiti Brunei Darussalam\\Brunei Darussalam}}
\email{nagender.aneja@ubd.edu.bn}


\begin{abstract}
Machine translation has many applications such as news translation, email translation, official letter translation etc. Commercial translators, e.g. Google Translation lags in regional vocabulary and are unable to learn the bilingual text in the source and target languages within the input. In this paper, a regional vocabulary-based application-oriented Neural Machine Translation (NMT) model is proposed over the data set of emails used at the University for communication over a period of three years. A state-of-the-art Sequence-to-Sequence Neural Network for  \ac{Malay to English} and \ac{English to Malay} translations is compared with Google Translate using Gated Recurrent Unit Recurrent Neural Network machine translation model with attention decoder. The low BLEU score of Google Translation in comparison to our model indicates that the application based regional models are better. The low BLEU score of \ac{English to Malay} of our model and Google Translation indicates that the Malay Language has complex language features corresponding to English.
  
\end{abstract}

%



\maketitle


\section{Introduction}
Machine translation is important for news translation, a biomedical translation, automatic post-editing task, chatbots for understanding different languages, and question/answer systems\cite{yu2019inferential}.  \ac{MT} provides metrics to assess (i) the translation quality given reference translation, (ii) the translation quality without access to any reference, (iii) a robust translation task, and (iv) a parallel comparable corpora task to improve the translations by parallelizing model for translation and searching the web for translation.  \ac{MT} quality have increased considerably with most notably advances in the field of \acf{NMT} (\cite{bulte2019neural}) by learning the mapping between source and target language via neural networks and attention mechanisms. 

Neural \ac{MT} models - \acl{RNN}, \acl{LSTM}, \acl{GRU}, Transformer (16-layer) ,  Transformer-Big (more hidden units)  \cite{wang2019learning, vaswani2017attention} are used for translation. These models are also used for selection and preparation of training data using comparable corpora for \ac{MT}.
The \ac{RNN} units stacked with 1-2 layers are sufficient for a small data set system which may be used for mobile applications or embedded system. However recently the system based on multi-layer self- attention has shown some improvement on large scale state-of-art datasets.

\cite{domingo2019demonstration} presented online learning for \ac{NMT} wherein authors integrated machine translation with the user interface so that machine continuously learn from human choices and adapt the model to a specific domain. \cite{yu2019inferential} presented a context-aware model for machine comprehension using an encoder, decoder and reinforcement learning.

In this paper, an application-based corpus populated with regional vocabulary, human translations and
corresponding translations of the email content from Google Translate is prepared for developing the neural machine translation model. We want to show that these types of models are required in comparison to commercial general translators e.g. Google translator. Therefore, a RNN based \acl{GRU} with attention decoder model is used for the University Email application, which predicts the next word conditioned on the previous context words \cite{zhang2019bridging}. The bilingual emails collected at the University for communication
over a period of three years in size is small in comparison with state-of-the-art-dataset e.g. WMT-18 (English $\rightarrow$ German).

The problem is found to require the context of the email content to be preserved during training on the dataset that may have multiple contexts. The problem has different challenges for  \ME and \EM translations.
The model developed for the problem initially was unable to learn the context for source and target languages within an email even in the presence of attention mechanism. Thus, the problem
needs more efforts and a different approach when the dataset has multiple contexts. The bilingual emails are compared with \ME Google translations and \EM Google translations, respectively. 

Table \ref{tab:EmailDataformat} depicts the format of the email corpus for the problem undertaken in the research. The trained model output sentence usually has multiple reasonable translations even if it generates a word different from ground truth word.\\
reference: Dear All
\\
candidate 1: All in all
\\
candidate 2: Dear all
\\
candidate 3: Respected all
\\
For example, the translation candidate 1 can be treated as a potential error in comparison to candidate 2 and candidate 3.

We observed that splitting the input email
based on the context before feeding it into RNN Encoder improved the performance over Google Translate by 10-20 BLEU points which means the model error was improved. However, we could not address all the problems observed in Google Translate. 


We improved the BLEU score over the regional vocabulary keeping the size of dataset small, including multiple contexts in an email. Results show that the training can be improved on the application scale, even with a small dataset and using a simple model rather than a very deep model. The results indicate that application-based regional models are better.

Our contributions are following
\begin{enumerate}
	\item Bilingual regional vocabulary populated email corpus with corresponding translations from Google Translate
	\item Trained \ac{NMT} model with higher BLEU score than Google Translate  
	\item Context-based results of language translations when using \ac{NMT} model
	\item Regional language Malay based application 
\end{enumerate}

\vspace{-10.0pt}
\begin{table}[H]
	\caption{Email Corpus format}    
	\begin{center}
		\begin{tabular}{ |p{1.3cm}|p{1.6cm}|p{1.9cm}|p{1.8cm}|} 
			\hline
			{ \bf eng human } & { \bf  malay human} & { \bf malay google translate } & { \bf eng google translate } \\ \hline
			Dear Students  & Pelajar yang dihormati &  Pelajar yang dihormati &  Dear student\\ 
			\hline
		\end{tabular}
		\label{tab:EmailDataformat}
	\end{center}        
\end{table}

\section{RNN-based NMT Model}
The approach can be used in applications of NMT models. We take the RNN-based NMT model to explain the method we used in this application \cite{bahdanau2014neural}. Assume the source email and observed translation email can be expressed as sequence of word as $x={x_1,x_2, \ldots, x_{|x|} }$ and $y={y_1,y_2, \ldots, y_{|y|} }$ respectively. 
The core of the \ac{NMT} is composed of sequence to sequence model generating translations using sequence to sequence or encoder-decoder network. The network consists of mainly three parts:

\begin{enumerate}
	\item Encoder
	\item Attention Context Vector
	\item Decoder
\end{enumerate}

{\bf Encoder:} An encoder is a stack of many recurrent units where each accepts a single word or element of the input sequence, process the element and pass the state forward. The hidden states $h_t$ are computed as in Equation \ref{eqn:encoder} with the help of current input $x_{t}$, previous  state $h_{t-1}$, and weights of the network $W$. This is the final hidden state of the encoder that is represented by Equation \ref{eqn:encoder}.

\vspace{-10.0pt}
\begin{equation}
h_t = f(W^{(hh)} h_{t-1}~+~W^{(hx)} x_t)
\label{eqn:encoder}
\end{equation}

{\bf Attention:} The context vector aims to encapsulate input sequence information to assist the prediction of another sequence by a decoder. This acts as an initial hidden state for the decoder. The context vector $c_p$ are computed as in Equation \ref{eqn:normalizedrelevenceita}, \ref{eqn:normalizedrelevenceitb} and \ref{eqn:contextvector} with the help of previous hidden state $h_{t-1}$, previous state $s_{p-1}$, and weights of the network normalized over the source sequence in Equation \ref{eqn:normalizedrelevenceitb} 

\vspace{-10.0pt}
\begin{equation}
r_{rp} = v^{T}_{a}tanh(W^{(ss)}{s_{p-1}}+ W^{(hh)}h_{t-1})
\label{eqn:normalizedrelevenceita}
\end{equation}

\vspace{-10.0pt}
\begin{equation}
\alpha_{tp} = \frac {exp(r_{tp})}{\Sigma^{|x|}_{t=1}exp(r_{tp})}  
\label{eqn:normalizedrelevenceitb}
\end{equation}

The source context vector $c_{p}$ is weighted sum of all source annotations and can be calculated in Equation \ref{eqn:contextvector}
\vspace{-5.0pt}
\begin{equation}
c_{p} = \Sigma^{|x|}_{t=1} {\alpha_{tp}}h_{t}
\label{eqn:contextvector}
\end{equation}

{\bf Decoder: } A decoder is similar to the encoder as it comprises of many recurrent units cells wherein each cell predicts an output word at a time step. Each recurrent unit cell accepts the previous target state $y_{p-1}$ and source context vector $c_p$ to produce output and next  target hidden state represented by Equation \ref{eqn:decoder}.

\vspace{-10.0pt}
\begin{equation}
s_p = f(W^{(ss)}. s_{p-1} + ~W^{(sy)} y_{p-1} + ~W^{(sc)} c_p)
\label{eqn:decoder}
\end{equation}

The $j^{th}$ target hidden state in the decoder is computed using the previous hidden state. The Probability distribution $P_j$ over all words in target vocabulary is produced from the decoder at any time, is computed using Softmax using Equation \ref{eqn:softmax}.

\vspace{-10.0pt}
\begin{equation}
t_j = f(W^{(ss)}. s_{j} + ~W^{(sy)} y_{j-1} + ~W^{(sc)} c_j)
\label{eqn:fc}
\end{equation}

\vspace{-10.0pt}
\begin{equation}
P_j = softmax(W^s . t_j)
\label{eqn:softmax}
\end{equation}

Thus, sequence to sequence model can map sequences of varying lengths to each other. 

\section{Approach: Neural Machine Translation model for email application with attention decoder}

\begin{figure*}[!htb]
	\centering
	\begin{subfigure}[b]{0.497901\textwidth}
		\centering
		\includegraphics[width=\textwidth]{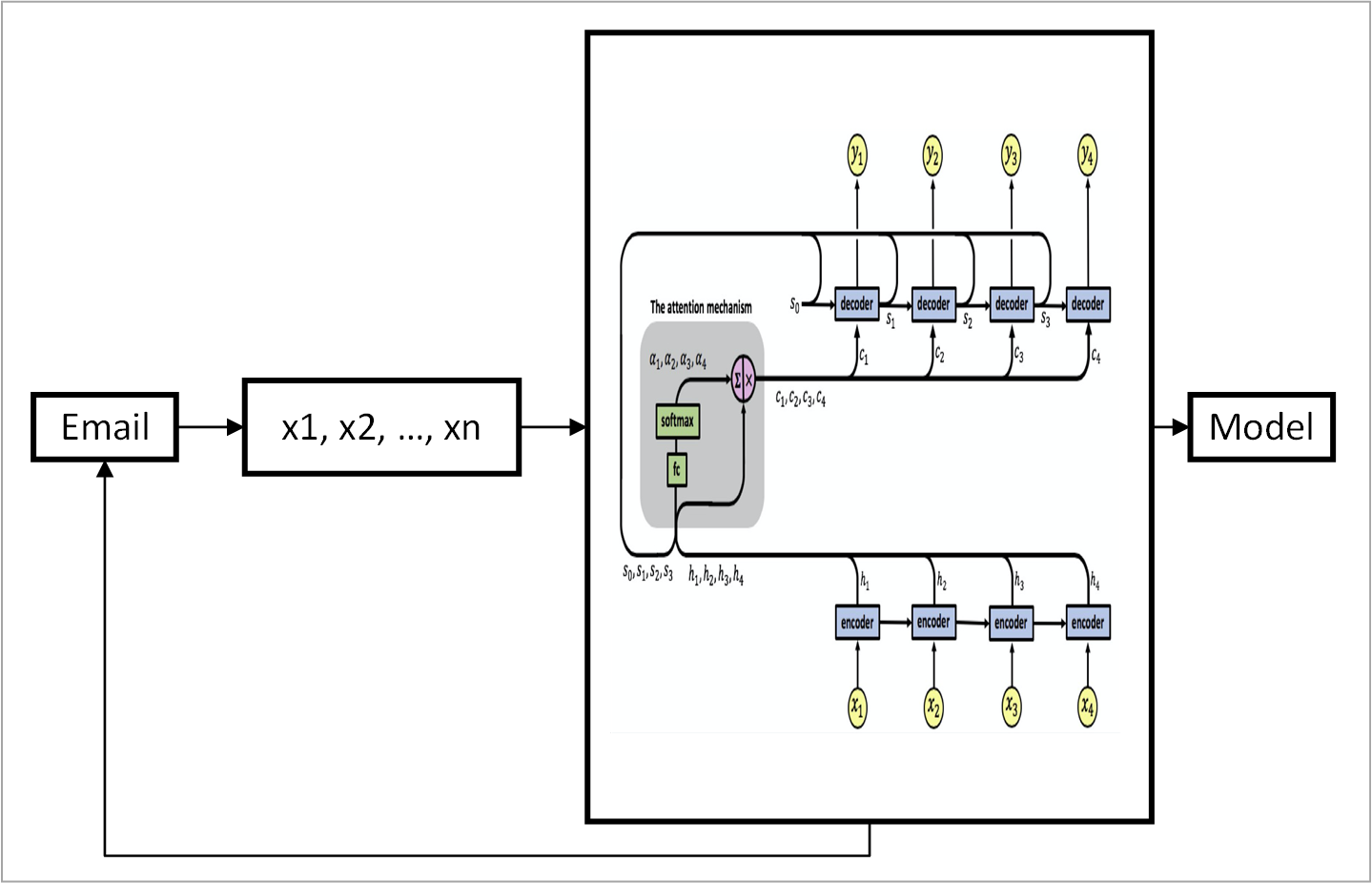}
		\caption{}    
		\vskip\baselineskip
		\label{fig:process1}
	\end{subfigure}
	\hfill
	\begin{subfigure}[b]{0.497901\textwidth}  
		\centering
		\includegraphics[width=\textwidth]{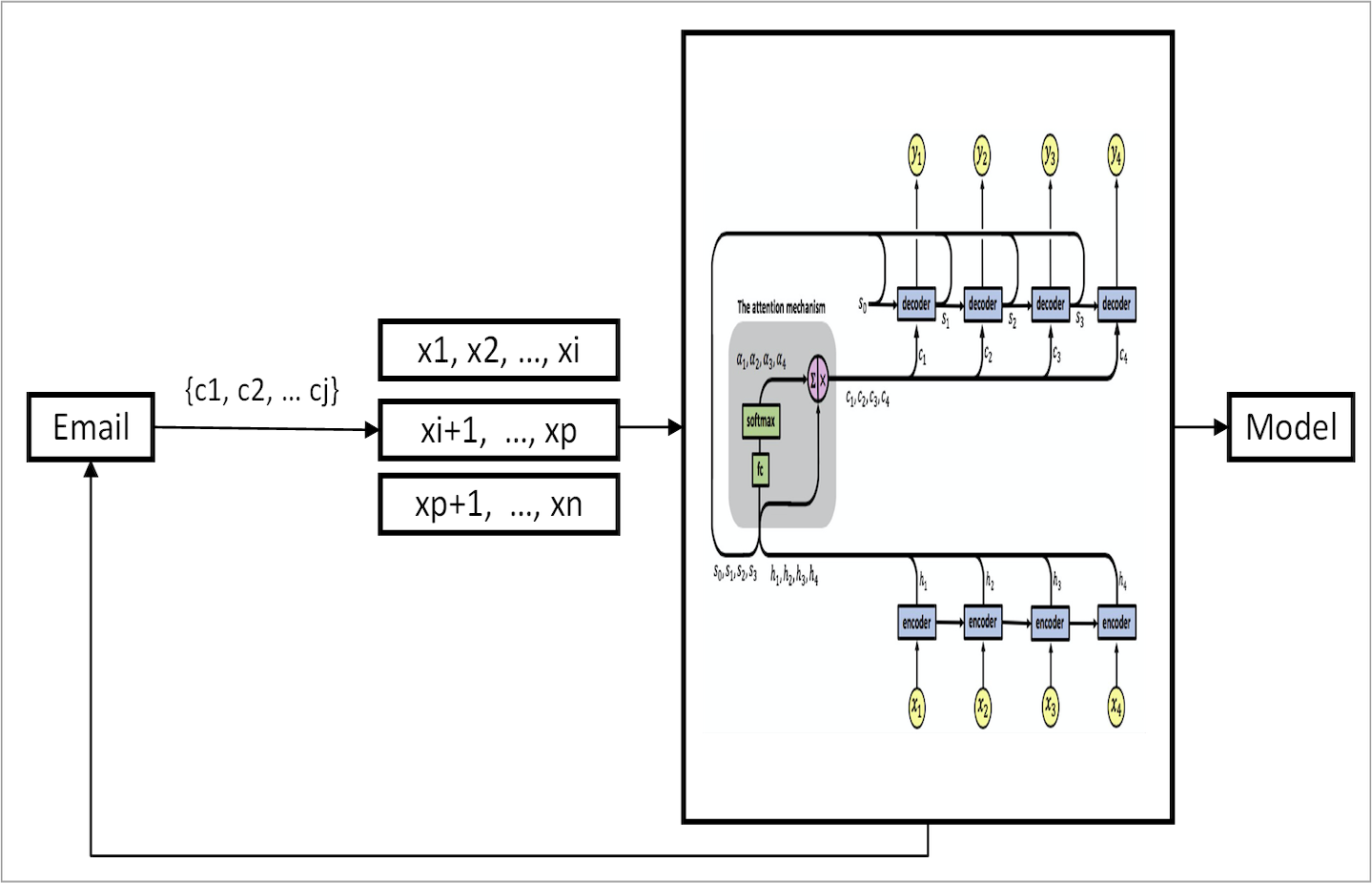}
		\caption{}
		\vskip\baselineskip
		\label{fig:process2}
	\end{subfigure}
	\hfill
	\caption{Neural Machine Translation model for email application (a) Solely-Email-Level Selection (b) Contextual-Paragraph-Level Selection}
	\label{fig:model}
\end{figure*}


The main aim of our method is to feed the higher context, i.e. splitting the input text into contextual content to increase the model output probability distribution so that it matches with the probability distribution of the ground truth values. This potentially can reduce the gap between training and inference by training the model to handle the situation, which will appear during test time. 

We discuss two methods to use NMT for the applications. To predict the p-th target word $y_{p}$, the following steps are involved in our approach:

\begin{enumerate}
	\item Solely-Email-Level Selection
	\item Contextual-Paragraph-Level Selection
\end{enumerate}

\subsection{Solely-Email-Level Selection} 
Solely-Email-Level Selection approach uses the full email that has multiple contexts in different paragraphs.

\vspace{-5.0pt}
\begin{equation}
r_{rp} = v^{T}_{a}tanh(W^{(ss)}{s_{p-1}}+ W^{(hh)}h_{t-1})
\label{eqn:normalizedrelevenceite}
\end{equation}
\vspace{-10.0pt}
\begin{equation}
\alpha_{tp} = \frac {exp(r_{tp})}{\Sigma^{|x|}_{t=1}exp(r_{tp})}  
\label{eqn:normalizedrelevenceitf}
\end{equation}

The source context vector is a weighted sum of all source annotations and can be calculated as in Equation \ref{eqn:contextvectorg}

\vspace{-5.0pt}
\begin{equation}
c_{p} = \Sigma^{|x|}_{t=1} {\alpha_{tp}}h_{t}
\label{eqn:contextvectorg}
\end{equation}

At the p-th step, the NMT model needs the ground truth word $y_{p-1}$ word as the context word to predict $y_{p}$, thus we need to select a $y_{p}$ which need to be similar to the ground truth word. We used Equations \ref{eqn:normalizedrelevenceite}, \ref{eqn:normalizedrelevenceitf} and \ref{eqn:contextvectorg}, however, the model could not converge. We used GRU model for translation, and it could not converge because Equations \ref{eqn:normalizedrelevenceite}, \ref{eqn:normalizedrelevenceitf} and \ref{eqn:contextvectorg} determine the context over the whole length of input text while the context may not be preserved in the different paragraphs of the input text. 

\subsection{Contextual-Paragraph-Level Selection} 
One option is to optimize the length of the input sequence such that the context is preserved in the input text. 

Assume the source email text with sequence of words $x = \{x_1, x_2, \ldots, x_{|x|} \}$ has multiple contextual paragraphs with context vector $c = \{c_1, c_2, \ldots, c_j\}$. The objective of this approach is to split the text vector comprising the words so that split vector is based on the context vector.

\vspace{-2.0pt}
$x={x_1, x_2, \ldots, x_{|c1|},}$

$ {x_{|c1|+1},x_{|c1|+2}, \ldots, x_{|c1|+|c2|}, }$ 

$ { x_{|c1|+|c2|+1},x_{|c1|+|c2|+2}, \ldots, x_{|c1|+|c2|+|c3|} }$

$ { x_{|c1|+|c2|+|c3|+1},x_{|c1|+|c2|+|c3|+2}, \ldots, x_{|c1|+|c2|+|c3|} }$

$\ldots$

${x_{|c1|+|c2|+|cj|+1} , x_{|c1|+|c2|+|cj|+2}, \ldots x_{|x|}} $

Equation \ref{eqn:contextvectorg} would be transformed to Equations \ref{eqn:contextvectorc}, \ref{eqn:contextvectord},\ref{eqn:contextvectore} according to the context vector.

\vspace{-5.0pt}
\begin{equation}
c_{1}{p} = \Sigma^{|c_{1}|}_{t=1} {\alpha_{tp}}h_{t} \\
\label{eqn:contextvectorc}
\end{equation}
\vspace{-10.0pt}
\begin{equation}
c_{2}{p} = \Sigma^{|c_{2}|}_{t=1} {\alpha_{tp}}h_{t} \\
\label{eqn:contextvectord}
\end{equation}
\vspace{-10.0pt}
\begin{equation}
c_{j}{p} = \Sigma^{|c_{j}|}_{t=1} {\alpha_{tp}}h_{t}
\label{eqn:contextvectore}
\end{equation}

The NMT model needs the ground truth word $y_{p-1}$ word as the context word to predict $y_{p}$, which is provided since context is maintained in the paragraph in Equations \ref{eqn:contextvectorc}, \ref{eqn:contextvectord}, \ref{eqn:contextvectore}.

Thus, at the p-th step the model converges. A comparison of Equation \ref{eqn:contextvectorg} and Equations \ref{eqn:contextvectorc}, \ref{eqn:contextvectord},\ref{eqn:contextvectore} clearly depicts the convergence of two scenarios and further explained in the Section \ref{ContextualVariation}. Figure \ref{fig:process2} 
demonstrates the proposed implementation of the model  for an application to optimize in comparison to implementation in Figure \ref{fig:process1}.

\subsection{Results and Analysis} \label{ResultsAndAnalysis}
The objective of this research is to develop an NMT model for English Malay emails which were circulated to Universiti Brunei Darussalam teaching staff and students. The international staff and students are part of the University and therefore the communication in many emails is in both the languages. 

We also paired each email with a translation from Google Translate. A sample of the dataset is shown in Table \ref{tab:EmailDataformat}. The table has four columns, the content of the first and second columns are from the email used in the communication. The third column is the English translation of email content in Malay language using Google Translate. The fourth column is the Malay translation of email content in English Language using Google Translate.

There are a few issues in Google Translate as shown in the table when translating English language \lq Dear All \rq to Malay Language, Google still shows \lq Dear All \rq. Additionally, one can see in other cases that the English and Malay translation from Google doesn't correspond to what has been used in the email. The problem is defined in terms of why the dataset of using emails is important. Since the translation of a sentence word-by-word is not that a user wants. In many cases, translation varies from person to person since, in many sentences or words, there are different ways to communicate the same matter.

\subsection{Experiment Settings}
We split 131 emails into contextual paragraphs. Thus, our dataset for English $\rightarrow$ Malay translation and vice versa comprises 785 contextual paragraph pairs. 

We divided the email into contextual paragraphs since training a model with full email which has multiple contextual paragraphs is error-prone. We also observed this fact when training the model with full email. This indicates that  \acl{GRU} \acl{RNN} machine translation model does not perform well in learning the patterns for long context, which in fact is not there, but the model tries to find and thus make an error.

The dataset dictionary comprises 1808 English words and 1628 Malay words. We set a maximum length of input text as 2000 to cover most sentences or paragraphs in the training process and also to terminate the output.  

To train the network, we pass the input sequence through the encoder and track every output and hidden state. Thus, after passing an input sequence with the initial hidden state, we get encoder outputs and the final hidden state. Passing an input sequence of five words with 256 hidden sizes will produce encoder outputs of the tensor size (5, 256) and final hidden state of 256 size tensor vector.

The decoder is then given the first input as $\langle$ SOS $\rangle$ token and final hidden state of the encoder. The decoder can be given next input as the best guess by the decoder or the real target outputs during the training process. The concept of using target outputs as the next input is called teacher forcing that helps to converge the training process faster. We used the teacher forcing algorithm randomly with a probability of 0.5 \cite{lamb2016professor}. However, during testing or evaluation time, the decoder is given the next input as the best guess only.

Network loss is computed based on decoder output and target tensor. Network weights are optimized using Stochastic Gradient Descent (SGD) optimizer using the initial learning rate of 0.01.  We stored loss after every 100 steps to track if the network is learning.

\begin{figure*}[!htb]
	\centering
	\begin{subfigure}[b]{0.24\textwidth}
		\centering
		\includegraphics[width=\textwidth]{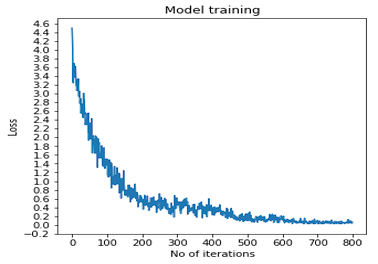}
		\caption{Malay $\rightarrow $ English }    
		\vskip\baselineskip
		\label{fig:MEloss}
	\end{subfigure}
	\hfill
	\begin{subfigure}[b]{0.24\textwidth}  
		\centering
		\includegraphics[width=\textwidth]{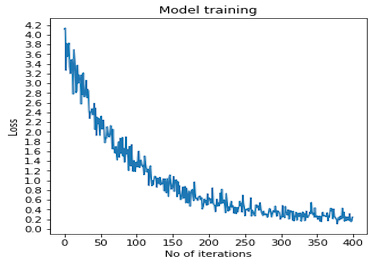}
		\caption{English $\rightarrow $ Malay }
		\vskip\baselineskip
		\label{fig:EMloss}
	\end{subfigure}
	\hfill
	\begin{subfigure}[b]{0.24\textwidth}
		\centering
		\includegraphics[width=\textwidth]{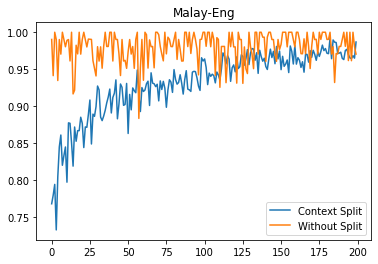}
		\caption{Malay $\rightarrow $ English }
		\vskip\baselineskip
		\label{fig:aw1}
	\end{subfigure}
	\hfill
	\begin{subfigure}[b]{0.24\textwidth}  
		\centering
		\includegraphics[width=\textwidth]{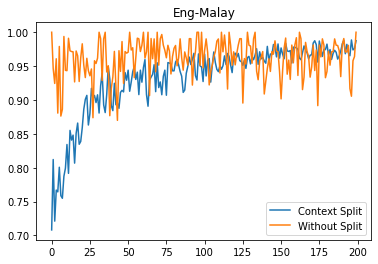}
		\caption{English $\rightarrow $ Malay }
		\vskip\baselineskip
		\label{fig:aw2}
	\end{subfigure}
	\hfill
	
	\caption{(a)-(b) Variation of model training (Negative Log Likelihood Loss Model Loss) x 100 iteration. Figure 2: (c)-(d) Variation of model context  x 100 iteration}
	\label{fig:aw}
\end{figure*}

\subsection{Model convergence}
In this section, we provide results for both translation models, namely Malay $\rightarrow $ English and English $\rightarrow $ Malay when we split the email with contextual paragraphs. We show results in terms of NLL loss, BLEU score \cite{papineni2002bleu} and comparison with Google Translation.  

The loss explains the training process so that how the training performed during the number of iterations. Figure \ref{fig:MEloss} shows graph of NLL loss with the number of iterations for Malay $\rightarrow $ English. We performed 80,000 iterations that reduced the cross-entropy loss from 4.498 to 0.023. Figure \ref{fig:EMloss} shows a graph of NLL loss with the number of iterations for English $\rightarrow $ Malay. We performed 40,000 iterations that reduced the loss from 4.14 to 0.106. These graphs show that the model training is computationally faster in English $\rightarrow $ Malay rather than Malay $\rightarrow $ English. 

\subsection{Contextual Variation} \label{ContextualVariation}
Figure \ref{fig:aw1} - \ref{fig:aw2} shows graphs with the number of iterations for the variation of context with English $\rightarrow $ Malay and  Malay $\rightarrow $ English wherein we observed that the context does not converge when email was not split in both Figure \ref{fig:aw1}  -\ref{fig:aw2} while, when the email split, context starts with a lower value and converge to 1 with an increase in the number of iterations. The context in the decoder depends on the attention weights calculated at each iteration. In a paragraph with multiple contexts, the attention weights are computed with a higher value and remain the same while the model does not converge. In the case of the contextual paragraph with split context, the attention weights start with a lower value and converge to 1, therefore, contributing to target predicted sequence to near the ground-truth value.

\subsection{BLEU Score Comparison}

Table \ref{table:bleuEMME} shows the comparison of BLEU score between (i) Malay  $\rightarrow $ Model English, (ii) English $\rightarrow $ Model Malay, (iii) Malay  $\rightarrow $ Google English  translation, and (iv) English $\rightarrow $ Google Malay.

We present BLEU score for randomly chosen 100 paragraphs from the dataset after the model is trained when the NLL Loss is negligible.

The low BLEU of English to Malay of our model and Google Translation indicates that the Malay Language has complex language features corresponding to English.

The low BLEU of Google Translation in comparison to our model indicates that the application based regional models with contextual split are better. 
\vspace{-20.0pt}
\begin{table}[!htbp]
	\caption{BLEU Score}
	\begin{tabular}{|l|l|l|l|}
		\hline
		Model             & Human Model &  Model             & Google Model\\ \hline
		E-\textgreater{}M & 0.95                 & E-\textgreater{}M & 0.75                 \\ \hline
		M-\textgreater{}E & 0.93                 & M-\textgreater{}E & 0.735        \\       \hline
	\end{tabular}
	\label{table:bleuEMME}
\end{table}


Below we present a sample output from the model. The sample shows input text, the true value expected from model and google, predicted text from model and predicted text from Google. \\

{\bf Input: } dengan itu para pensyarah pegawai kakitangan dan para pelajar dan juga alumni universiti brunei darussalam adalah amat dialu alukan untuk turut serta menjadi pembimbing bagi program ini. \\
{ \bf Truth: } therefore lecturers officers staff and students as well as university brunei darussalam alumni are welcome to participate in this program . \\
{ \bf Pred: } therefore lecturers officers staff and students as well as university brunei darussalam alumni are welcome to participate in this program . \\
{ \bf Google: } therefore the faculty staff staff and students as well as the university of brunei darussalam alumni are welcome to participate in the program . \\

\section{Related Work}

Researchers \cite{sutskever2014sequence} have used five deep layered \acl{LSTM} training model improving the existing result of WMT-14 dataset for an \acl{English to French} translation of fixed conditionality. Cho et al.  \cite{cho2014learning} improved hidden unit of LSTM \acl{RNN} by dropping a previously hidden unit whenever there is irrelevant information.  The reset gate and update gate collectively improve the hidden state. The states which capture short term dependencies were including reset gate while long term dependencies were captured with updated gate. The research also incorporated phrase pair based dependencies to improve the model. 

The authors in \cite{ruzsics2017neural} improved the \acl{LSTM} \acl{RNN} model by integrating it with canonical segmentation of words by providing the exact words of the verb. This is an improvement over the character level for language model with morphemes. The length constraint due to segmentation is also included in the model to handle variable-length sequences.

\acl{MT} has been promising but on state-of-art datasets research is still in progress by reducing the noise between source and target sentence-level, reducing the overcorrection at word-level as well as at sentence level. The authors \cite{zhang2019bridging} proposed to solve overcorrection problem by selecting predicted word as next input rather than ground truth by defining a measure on BLEU score on word-level as well as on sentence-level. 

The recent problem under application of \acl{MT} is inferential machine comprehension. The inferential network \cite{yu2019inferential} is proposed to comprise a micro infer cell where  one master unit is for reading the document to locate the ending of reasoning for the question according to context of the question. The reader unit uses attention mechanism from the reasoning operation to retrieve the content. The writer unit write the content to memory cell. The problem can not be optimized using back propagation therefore reinforcement learning is used to terminate the mechanism. Another approach in the direction of improvement of \acl{NMT} is to augmenting the source sentences \cite{bulte2019neural} with fuzzy matches from translation memory using similarity match.


\cite{papineni2002bleu} presented `BLEU' a metric for automatic evaluation of \ac{MT}. Human evaluation of a translation is exhaustive but expensive. The authors defined n-gram precision over a corpus by first comparing the n-grams of the reference sentence with n-grams of the candidate translations in the corpus. BLEU score is given higher value when the number of n-gram matches are higher. 


\cite{tan2017evaluating} used \ac{LSTM} for POS tagging over the Malay Language dataset. The authors compared the Weighted Finite-State Transducer, Hidden Markov Model (HMM) and \acl{LSTM}  for tagging and found the state transducer produced more accurate tagging with morphological information. 


The work \cite{ranaivo2007identifying} provided a rule-based method to identify unknown words in the corpus of Malay language. The abbreviations, affixed words, proper nouns, and loanwords were mainly identified and classified unknown words types.

\section{Conclusion and Future Work}
In this paper, the performance of \ac{RNN} based NMT model namely \ac{GRU} with attention decoder is presented on our dataset populated with English-Malay translated emails circulated at the University. The model was unable to learn the context for source and target language within the input text even in the presence of attention mechanism. Thus, a different approach splitting the input text into contextual content is used. General purpose trained model doesn’t perform well for a specific application. Thus, there is need to develop application oriented trained model populated with application specific vocabulary. The model using regional email vocabulary showed 10-20 BLUE score better than google translate. The model was unable to learn when source input contains bilingual text. Thus, there is need to update general translators for multilingual blended input text.

\bibliographystyle{acm}
\bibliography{mt_paper}

%
%
%
%
%
%
%
%
%
%
%
%
%

\end{document}